\documentclass{article} % For LaTeX2e
\usepackage{iclr2015,times}
\usepackage{url}
\usepackage{amsmath, bm}
\usepackage{amsfonts}
\usepackage{graphicx}
\usepackage{caption}
\usepackage{subcaption}
\usepackage{wrapfig}
\usepackage[utf8]{inputenc} % set input encoding (not needed with XeLaTeX)
\usepackage[titletoc,title]{appendix}

\newcommand{\norm}[1]{\left\lVert#1\right\rVert}
\newcommand{\Exp}{\mathbb{E}}
\newcommand{\Four}{\mathcal{F}}

\title{The local low-dimensionality\\of natural images}

\author{
Olivier J. Hénaff, Johannes Ballé, Neil C. Rabinowitz \& Eero P. Simoncelli \\ %\thanks{ Use footnote for providing further information about author (webpage, alternative address)---\emph{not} for acknowledging funding agencies. Funding acknowledgements go at the end of the paper.}
Howard Hughes Medical Institute, and\\
Center for Neural Science\\
New York University\\
New York, NY 10003, USA \\
\texttt{\{ojh221, jb4726, nr66, eero.simoncelli\}@nyu.edu} \\
}
\newcommand{\En}{E}

\iclrfinalcopy % Uncomment for camera-ready version

\iclrconference % Uncomment if submitted as conference paper instead of workshop

\DeclareMathOperator{\Cov}{Cov}
\DeclareMathOperator{\Ve}{Vec}

\begin{document}

\maketitle

\begin{abstract}
We develop a new statistical model for photographic images, in which the local responses of a bank of linear filters are described as jointly Gaussian, with zero mean and a covariance that varies slowly over spatial position. We optimize sets of filters so as to minimize the nuclear norm of matrices of their local activations (i.e., the sum of the singular values), thus encouraging a flexible form of sparsity that is not tied to any particular dictionary or coordinate system. Filters optimized according to this objective are oriented and band-pass, and their responses exhibit substantial local correlation. We show that images can be reconstructed nearly perfectly from estimates of the local filter response covariance alone, and with minimal degradation (either visual or MSE) from low-rank approximations of these covariances. As such, this representation holds much promise for use in applications such as denoising, compression, and texture representation, and may form a useful substrate for hierarchical decompositions.
\end{abstract}
\section{Introduction}
%\begin{wrapfigure}{r}{0.5\textwidth}
%  \begin{center}
%\vspace{-0.15\linewidth}
%    \includegraphics[width=0.48\textwidth]{pictures/fig2/globalXY}
%  \end{center}
%  \caption{Global responses of localized filters to natural images are heavy tailed. Left: joint samples and marginal histograms of two oriented band-pass filters (shown at top). Right: same, for phase-randomized versions of filters in left.}
%\label{global}
%\end{wrapfigure}
\begin{wrapfigure}{r}{0.31\textwidth}
  \begin{center}
\vspace{-0.15\linewidth}
    \includegraphics[width=0.3\textwidth]{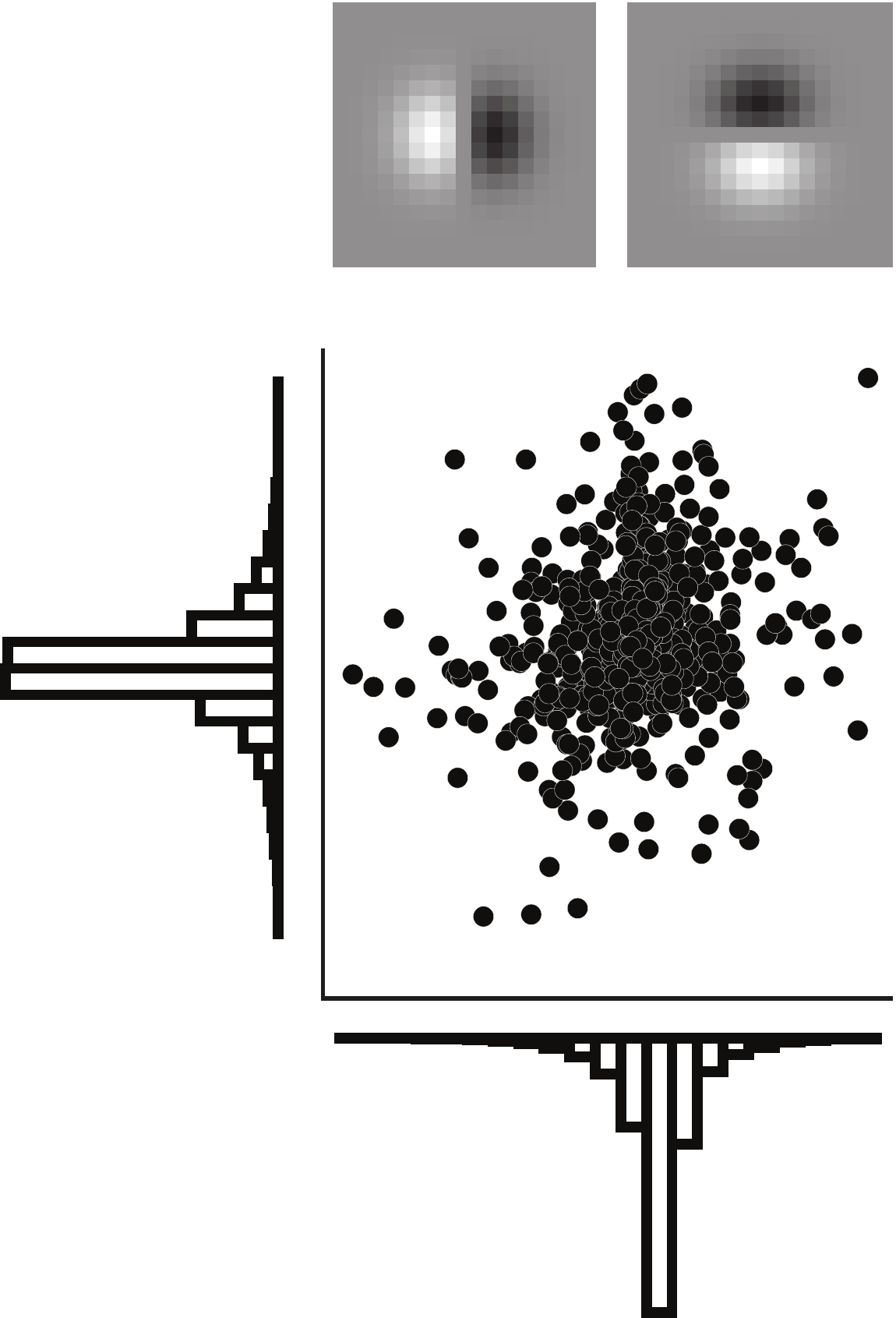}
  \end{center}
  \caption{Global responses of oriented band-pass filters to natural images are heavy tailed.}
\vspace{-0.15\linewidth}
\label{global}
\end{wrapfigure}

Many problems in computer vision and image processing can be formulated in terms of statistical inference, based on probabilistic models of natural, photographic images. Whereas natural images themselves are complex, locally structured, and high-dimensional, the vision community has traditionally sought probabilistic models of images that are simple, global, and low-dimensional. For example, in the classical spectral model, the coefficients of the Fourier transform are assumed independent and Gaussian, with variance falling with frequency; in block-based PCA, a set of orthogonal filters are used to decompose each block into components that are modeled as independent and Gaussian; and in ICA, filters are chosen so as to optimize for non-Gaussian (heavy-tailed, or ``sparse'') projections (\cite{Bell97}; figure~\ref{global}).

To add local structure to these models, a simple observation has proved very useful: the local variance in natural images fluctuates over space \citep{Ruderman:1994uu, Simoncelli:1997vm}. This has been made explicit in the Gaussian Scale Mixture model, which represents neighborhoods of individual filter coefficients as a single global Gaussian combined with a locally-varying multiplicative scale factor \citep{Wainwright99b}. The Mixture of GSMs model builds upon this by modeling the local density as a sum of such scale mixtures \citep{GuerreroColon:2008wh}.

Here, we extend this concept to a richer and more powerful model by making another simple observation about the local structure of natural images: the covariance of filter coefficients at a given location can vary smoothly with spatial position, and these local covariances tend to be highly elongated, i.e.\ they lie close to low-dimensional subspaces. In section~\ref{sec:analysis}, we motivate the model, showing that these properties hold for a pair of simple oriented filters. We find that the low-dimensionality of local covariances depends on both the filter choice, and the content of the images---specifically, it does not hold for phase-randomized filters or images. 
%either random filters or phase-randomized images. 
In section~\ref{sec:learning}, we use this distinctive property to optimize a set of filters for a measure of low-dimensionality over natural images. Finally, in section~\ref{sec:synthesis}, we demonstrate that the local low-dimensional covariance description captures most of the visual appearance of images, by synthesizing images with matching local covariance structure.

\section{Analysing local image statistics}
\label{sec:analysis}
\begin{figure}[h!]
\includegraphics[width=\textwidth]{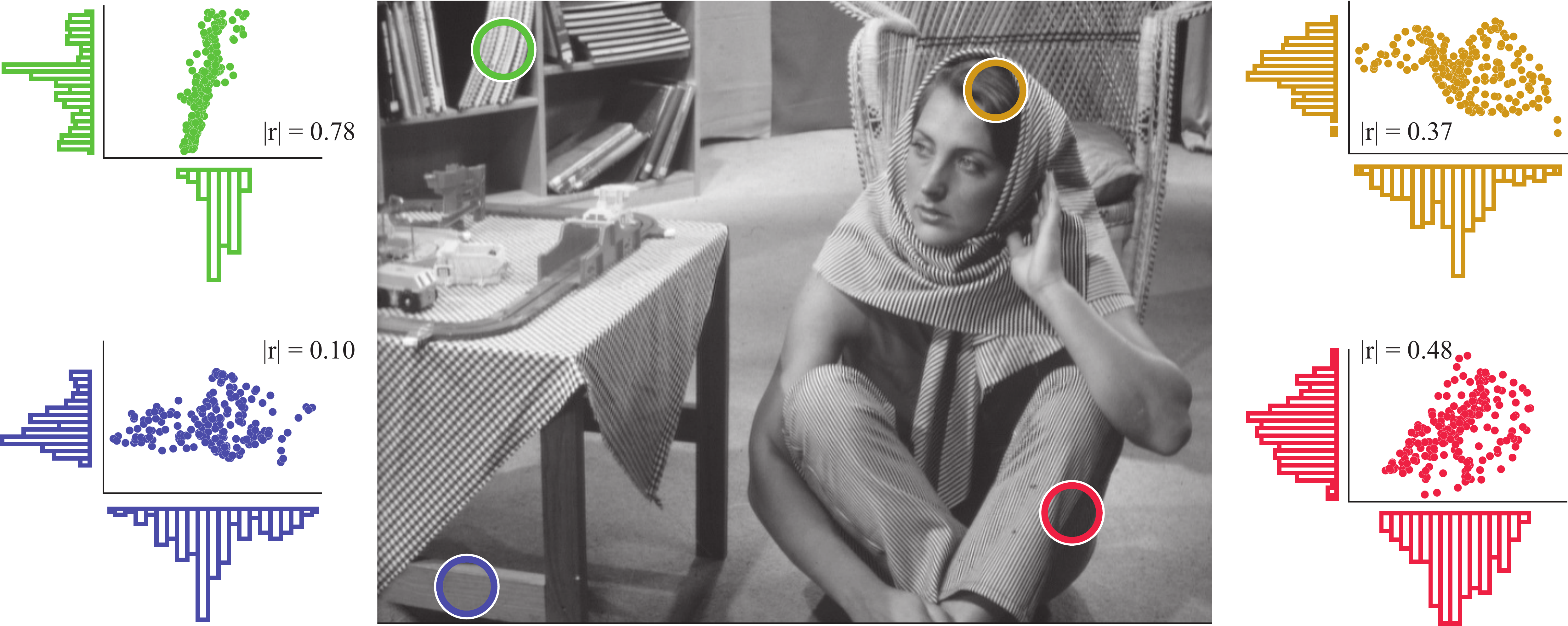}
\caption{Locally, we can approximate the responses of a pair of oriented band-pass filters to photographic images as jointly Gaussian, with a covariance that changes continuously across space. In regions with oriented content, these responses are low-dimensional, as indicated by a high correlation between filter responses.}
\label{scatter}
\end{figure}
To understand the statistics of local image patches, we begin with a simple example, based on analysis with two orthogonal, oriented band-pass filters. If we aggregate the two filters' responses over the whole image, the resulting joint distribution is approximately spherically symmetric but the marginal distributions are heavy-tailed (\cite{Zetzsche99,Lyu08c}; figure~\ref{global}). However, if we aggregate only over local neighborhoods, the distributions are more irregular, with covariance structure that varies in scale (contrast/energy), aspect ratio, and orientation (figure~\ref{scatter}). Our interest here is in the second property, which provides a measure of the dimensionality of the data. Specifically, a covariance with large aspect ratio (i.e., ratio of eigenvalues) indicates that the filter responses lie close to a one-dimensional subspace (line).

\begin{figure}
               \includegraphics[width=0.3\textwidth]{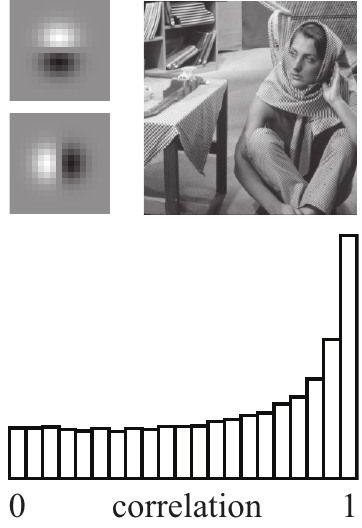}
        \hfill
                \includegraphics[width=0.3\textwidth]{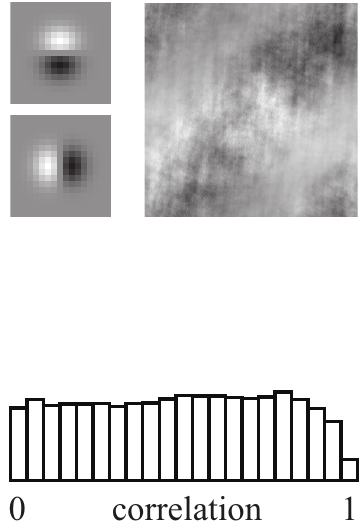}
        \hfill
                \includegraphics[width=0.3\textwidth]{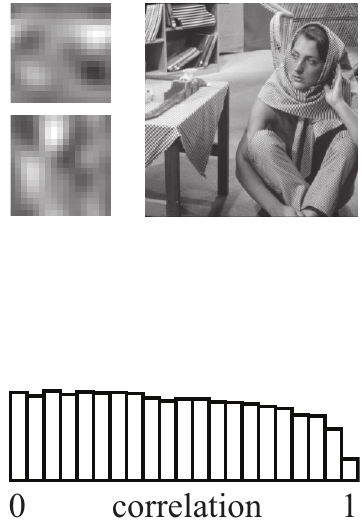}
\caption{Local low-dimensional structure is not a guaranteed property of all filter responses to all images. For each panel, the pair of filters in the top left corner (enlarged 15$\times$) are applied to the image in the top right, and the histogram of local correlation coefficient magnitudes across locations is plotted below. Oriented filters analyzing natural images (left) exhibit locally low-dimensional responses, but when oriented filters are applied to spectrally matched noise images (center) or phase-randomized filters are applied to photographic images (right) this behavior vanishes.}
\label{histogram}
\end{figure}

In the case of two filters, we can construct a simple, continuous measure of local dimensionality by calculating the correlation coefficient between filter responses in local neighborhoods. The distribution of correlation coefficient magnitudes across image patches is very skewed (figure~\ref{histogram}, left): in many locations, the responses are correlated, i.e.\ the local Gaussians are low-dimensional. In contrast, if we repeat the same experiment with spectrally-matched noise images rather than a photograph (figure~\ref{histogram}, center), the correlations are typically lower, i.e.\ the local Gaussians are more high-dimensional. The spectral properties of natural images alone are thus insufficient to produce local low-dimensional structure. Similarly, if we analyze a photograph with phase-randomized filters (figure~\ref{histogram}, right), we do not find the same local low-dimensionality. We take this as evidence that local low-dimensional structure is a property that emerges from the combination of local band-pass filters and photographic images.
\section{Local low dimensionality as a learning objective}
\label{sec:learning}
\subsection{Objective function}
We now ask whether these oriented band-pass filters are the best filters, or indeed the only filters, for producing representations of natural images that are locally low-dimensional. Just as marginal sparsity has been used as an objective for optimizing filters for representation of natural images \citep{Olshausen:1996uo, Bell97}, we aim to derive a filter bank that minimizes a measure of the local dimensionality of responses to natural images. Here we describe the construction of the objective function and the motivation behind its components; in section~\ref{sec:optimization} we cover some technical details relating to the optimization; in section~\ref{sec:results} we present the results.

To begin, suppose we have a convolutional filter bank $\bm f$, and an image $x$. We compute a map of filter responses $y_i (t) = (f_i * x)(t)$, with the index $t$ indicating spatial position, and consider this map in terms of a set of overlapping patches. %For each patch $P$, we have a collection of filter response vectors, which we can form into a matrix, $\bm Y_P = [\bm y(t) ]_{t \in P}$ composed of all the response vectors in that patch.
For each patch $P$, we can form a matrix $\bm Y_P = [\bm y(t) ]_{t \in P}$ composed of all the response vectors in that patch. 

Next, we need to define an appropriate measure of dimensionality for the local covariance on each patch. The correlation coefficient presented in section~\ref{sec:analysis} does not extend beyond the simple two-dimensional case. Instead, we choose to measure the nuclear norm of $\bm Y_P$, i.e.\ the sum of its singular values. This is the convex envelope of the matrix rank function, so it provides a continuous measure of dimensionality; unlike the rank, it is robust to small perturbations of filter responses away from true subspaces. The local low-dimensionality component of the objective is thus:
\[ \En_{local \ dim} = \sum\nolimits_P \norm{\bm Y_P }_*\] % = \sum\nolimits_P \sum\nolimits_i \sigma_i(\bm Y_P) where the $\sigma_i(\bm Y_P)$ are the singular values of $\bm Y_P$.
To ensure that the filter responses provide a reasonable representation of the original image, we reconstruct the image from them via the filter bank transpose $\bm\tilde{\bm f}(t) = \bm f(-t)$, and penalize for reconstruction error:
%of the linear transform corresponding to the filter bank $f$. This is another filter bank $\tilde{f}$ whose kernels are horizontally- and vertically-flipped versions of $f$. We then penalize any deviation of this reconstructed image $\sum_i z_i$ from the original one.
\[ \En_{recons} = \sum\nolimits_t \left( x(t) - \sum\nolimits_i  ( \tilde{f}_i * y_i ) (t) \right)^2\]
%\sum\nolimits_t \left( x - \sum\nolimits_i z_i \right)^2 =
%= \sum\nolimits_t \left( x - \sum\nolimits_i \tilde{f}_i * f_i * x \right)^2
Finally, in order to ensure that all filters are used and the filter responses are sufficiently diverse, we include a term that makes the global distribution of filter responses high-dimensional. One way to achieve this would be to form the matrix $\bm Y$ of all response vectors from the ensemble, and maximize its nuclear norm:
\[ \En_{global \ dim} = - \norm{ \bm Y }_* \]
Together with the reconstruction penalty, this tends to whiten the filter responses (i.e. $\Cov[ \bm y ] \propto \bm I$). %, see appendix~A
In practice however, this term allows degenerate solutions with filters that are related to each other by a phase shift. This is best understood in the Fourier domain: with the Parseval identity,
\begin{align*}
\Cov[ \bm y ]
& = \left[ \int \! y_i(t) y_j(t) \, \mathrm{d}t \right]_{i,j} \\
& = \left[ \int \! \hat{y}_i(\omega) \overline{\hat{y}_j(\omega)} \, \mathrm{d}\omega \right]_{i,j}\\
 & = \left[ \int \! |\hat{x}(\omega)|^2 \hat{f}_i(\omega) \overline{\hat{f}_j(\omega)} \, \mathrm{d}\omega \right]_{i,j} \\
\end{align*}
where $\hat{a} = \Four[ a ]$ is the Fourier transform of $a$.
% to the other. To remove the possibility of obtaining phase shifted copies of the same filter, we maximize the global dimensionality of filter reconstructions rather than activations. This improvement is better understood in the Fourier domain:
Two filters with identical Fourier magnitudes but different phases can make this expectation zero. To eliminate this degeneracy, we can maximize the dimensionality of the filter reconstructions $z_i = \tilde{f}_i * f_i * x$, rather than the filter responses $y_i = f_i * x$. As
\[ \Cov[ \bm z ] = \left[ \int \! |\hat{x}(\omega)|^2 |\hat{f}_i(\omega)|^2 |\hat{f}_j(\omega)|^2 \, \mathrm{d}\omega \right]_{i,j} \]
maximizing the nuclear norm of $\bm Z = [ \bm z(t) ]_t$ pushes $\Cov[\bm z ]$ towards a multiple of the identity and hence penalizes any overlap between filter Fourier magnitudes. Since this tends to insist that filters have hard edges in the Fourier domain, we relax this constraint by only penalizing for overlaps between blurred versions of the filters' Fourier magnitudes. Using a Gaussian blurring window $h$, we compute modulated filter reconstructions $\tilde{z}_i = (h \tilde{f}_i ) * (h f_i ) * x$, and whiten 
\[ \Cov[\bm\tilde{\bm z} ] = \left[ \int \! |\hat{x}(\omega)|^2 |h*\hat{f}_i(\omega)|^2 |h*\hat{f}_j(\omega)|^2 \, \mathrm{d}\omega \right]_{i,j} \]
by maximizing the dimensionality of $\bm\tilde{\bm Z} = [ \bm\tilde{\bm z}(t) ]_{t}$ via the term
\[ \En_{global \ dim} = - \norm{ \bm \tilde{\bm Z} }_* \]
Summarizing, we optimize our filter bank for
\begin{alignat*}{3}
\En & = \En_{local \ dim} && + \lambda \ \En_{recons} && + \mu \ \En_{global \ dim} \\
& = \sum\nolimits_P \norm{\bm Y_P }_* && + \lambda \norm{x - \sum\nolimits_i z_i }_2^2 && - \mu \norm{ \bm \tilde{\bm Z} }_*\\
\end{alignat*}
%where $\tilde{Z} = [ \tilde{z_i}(t) ]_{i,t}$ is a matrix composed of modulated filter reconstructions $\tilde{z}_i = (h \tilde{f}_i ) * (h f_i ) * x$ and $h$ is a Gaussian blurring kernel.
We now describe the practical details of the learning procedure and our results.

\begin{wrapfigure}{r}{0.5\textwidth}
  \begin{center}
\vspace{-0.2\linewidth}
    \includegraphics[width=0.48\textwidth]{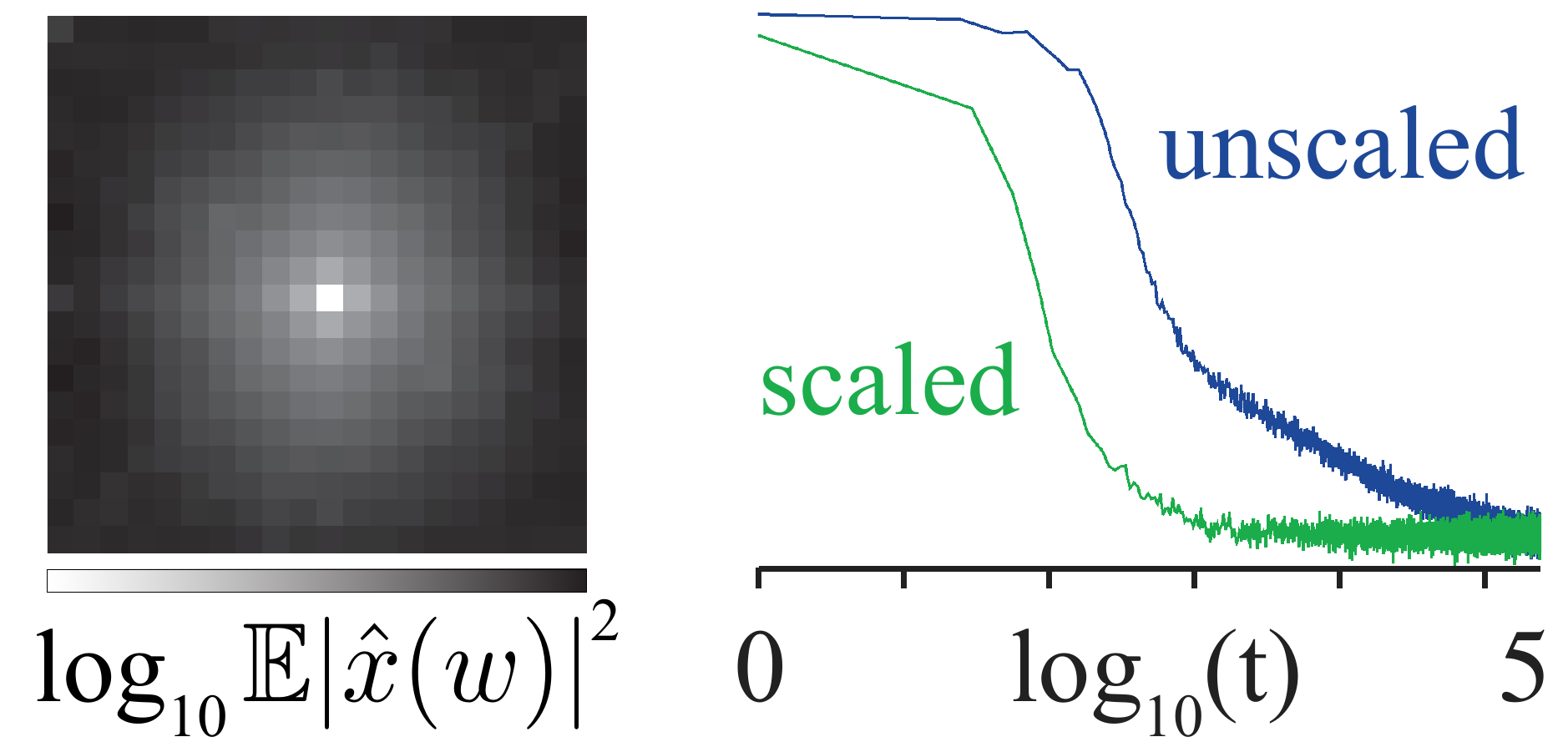}
  \end{center}
  \caption{Gradient scaling. Left: Input spectrum ranges from 1 to 100. Right: Value of objective over time, with and without gradient scaling.}
%\vspace{-0.1\linewidth}
%\vspace{-0.08\linewidth}
\label{scaling}
\end{wrapfigure}\subsection{Optimization}\label{sec:optimization}
\subsubsection{Model specifications}
We trained the model described above on the van Hateren dataset \citep{hateren_schaaf_1998} using the Torch machine learning library \citep{Collobert:2011wv}. We used $20\times 20$ pixel filter kernels, varying in number from 4 to 12, and estimated local dimensionality over neighborhoods of size $16\times 16$ pixels, weighted by a Gaussian window with a standard deviation of 3~pixels. We fixed the blurring window $h$ to be Gaussian with a standard deviation of 3~pixels, such that it only becomes negligible at the kernel boundary. The hyperparameters $\lambda$ and $\mu$ are easily tuned: we increased the reconstruction weight $\lambda$ until a desired reconstruction level was reached (e.g. a relative $L_2$ error of 1\%) and increased the diversity weight $\mu$ until none of the filters were zero nor perfectly correlated with another. In the experiments below they were set to 3500 and 100 respectively. We optimized our filter bank using stochastic gradient descent with a fixed learning rate, chosen as high as possible without causing any instability.
\subsubsection{Accelerated learning with gradient scaling} \label{sec:scaling}
We developed a method to scale gradients according to the input spectrum, and found that it considerably accelerated the optimization procedure. In ordinary gradient descent, the descent direction for the filter $f_i(t)$ is the negative gradient. Using the chain rule, it can be expressed in terms of the filter responses $y_i = f_i * x$:
\[ \Delta f_i(t) = - \frac{ \partial E}{\partial f_i(t)} = - \left( \frac{ \partial E}{\partial y_i} * \tilde{x}\right) (t) \]
In the Fourier domain, this is
\[ \Delta \hat{f}_i(\omega) = - \frac{ \partial E}{\partial \hat{f}_i(\omega)} = - \hat{x}(\omega) \frac{ \partial E}{\partial \hat{y}_i(\omega)} \] %(see appendix~B for details). 
Due to the hyperbolic spectrum of $\hat{x}(\omega)$ (figure~\ref{scaling}, left panel), the low frequency components of the gradient are substantially larger than the high frequency ones. We offset this imbalance by dividing the
gradient by the corresponding mean image frequency magnitude. The modified descent direction is thus
\[ \Delta \hat{f}_i(\omega) = \frac{-1}{\sqrt{\Exp \left[ |\hat{x}(w)|^2 \right]}} \frac{ \partial E}{\partial \hat{f}_i(\omega)} = \frac{ -\hat{x}(\omega) }{\sqrt{\Exp \left[ |\hat{x}(w)|^2 \right] }} \frac{ \partial E}{\partial \hat{y}_i(\omega)}\]
where the expectation is over the ensemble of all images.
% with two Fourier transforms
%\[ \Delta f_i(t) \leftarrow \Four^{-1}\left[ \frac{-1}{\sqrt{\Exp |\hat{x}(w)|^2}} \Four \left[ \frac{ \partial E}{\partial f_i}\right] \! (\omega) \right] (t)\]
%We estimate the mean frequency magnitude of the dataset once, then divide all gradients by this mean rather than by the local frequency magnitude $\hat{x}(\omega)$ in order to avoid any instability due to small values of $\hat{x}(\omega)$, and to reduce the complexity of the operation. We optimized two identical networks, with and without
The gradient scaling algorithm accelerates convergence by a factor of at least 10 (figure~\ref{scaling}, right panel) and does not affect the final result.
\subsection{Results}
\label{sec:results}
\begin{figure}[h!]
\includegraphics[width= 0.333\linewidth]{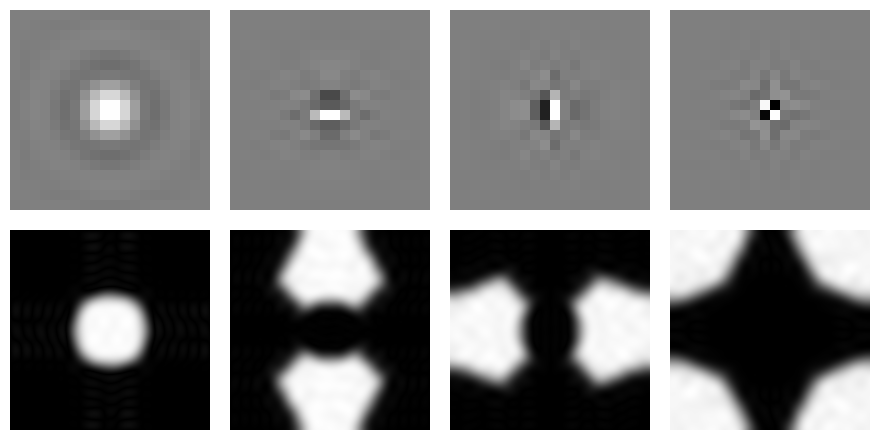}
\hfill
\includegraphics[width = 0.583333\linewidth]{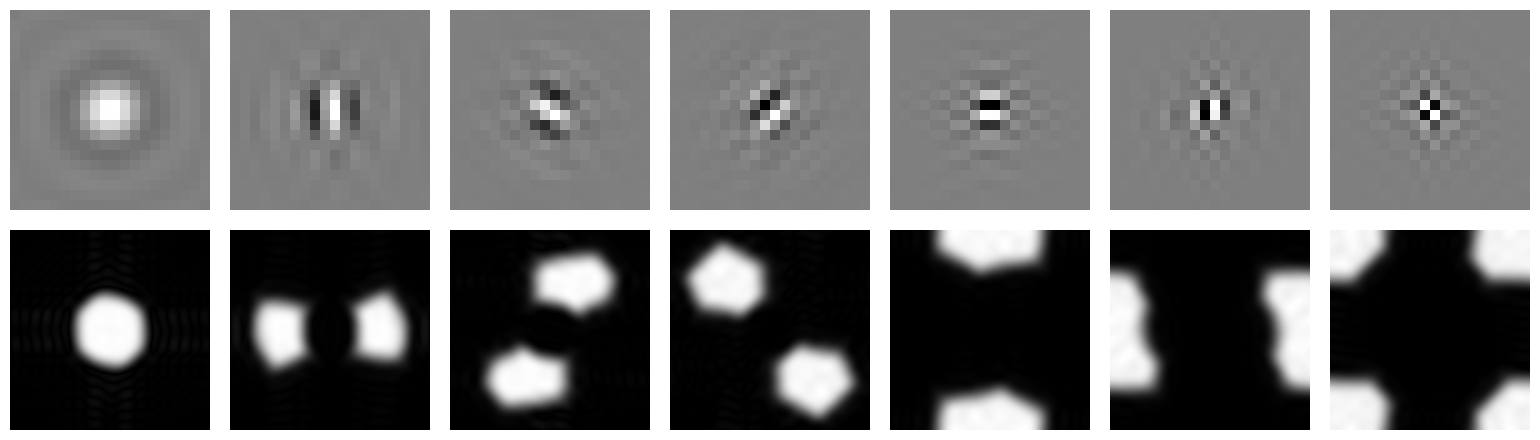}
\\
\vspace{0.02 \linewidth}
\\
\includegraphics[width = \linewidth]{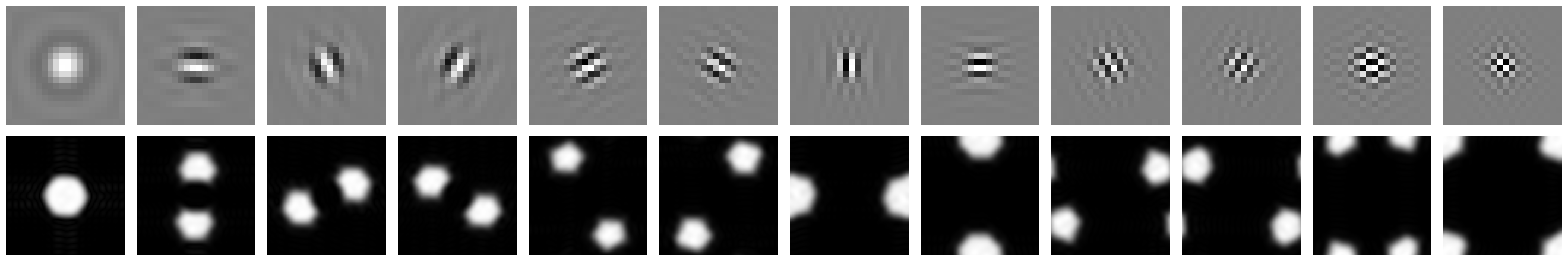}
\caption{Examples of optimized filter banks of size 4, 7 and 12. Top row: spatial filters, bottom row: Fourier magnitudes (zero frequency is in center).}
\label{filters}
\end{figure}
Even though our objective function is not convex with respect to the filter bank, we found empirically that different initializations lead to qualitatively similar results. The optimized filter bank for uncovering the local low-dimensional structure of natural images is composed of a low-pass, a high-pass, and a set of oriented band-pass filters (figure~\ref{filters}). As we increase the number of filters, we obtain a finer partition of scales and orientations: 7 filters divide the band-pass region into 2 radial sub-bands, with 3 orientations in the mid-low band and 2 orientations in the mid-high band. With 12 filters, the mid-low band remains mostly unchanged, while the mid-high band is partitioned into 6 orientations.
\section{Representing natural images as local covariance maps}
\label{sec:synthesis}
\begin{figure}[h!]
        \begin{subfigure}[b]{0.24\textwidth}
                \includegraphics[width=\textwidth]{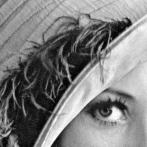} \\
%\vspace{0.94\linewidth}
\vspace{1.995\linewidth}
\caption{Original image \newline \newline \#pixels: 21600 \newline}
\label{vanillaA}
        \end{subfigure}
        \hfill
        \begin{subfigure}[b]{0.24\textwidth}
\includegraphics[width=\textwidth]{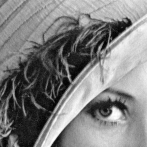} \\
\vspace{-0.07\linewidth}
\includegraphics[width=\textwidth]{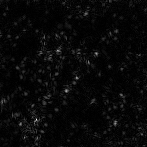}\\
\vspace{-0.07\linewidth}
\includegraphics[width=\textwidth]{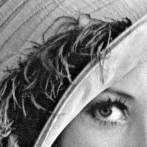}

%\caption{%relative error = 1.5\% \\
%		$\norm{x}_2 / \norm{\Delta x}_2$ = 1.5\% \\
%		neighborhood: 8$\times$8 \\
%		subsampling: 2$\times$2\\
%		\#measurements: 54000}
\caption{neighborhood: 8$\times$8 \\
		subsampling: 2$\times$2\\
		\#measurements: 54000 \\
		$\norm{\Delta x}_2 / \norm{x}_2$ = 1.5\%
		%relative error = 1.5\% \\		
}
\label{vanillaB}
        \end{subfigure}
        \hfill
        \begin{subfigure}[b]{0.24\textwidth}
\includegraphics[width=\textwidth]{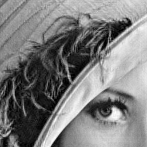} \\
\vspace{-0.07\linewidth}
\includegraphics[width=\textwidth]{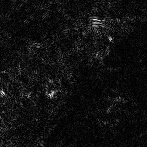} \\
\vspace{-0.07\linewidth}
\includegraphics[width=\textwidth]{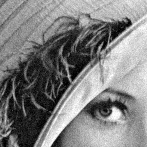}
\caption{
		neighborhood: 16$\times$16 \\
		subsampling: 4$\times$4\\
		\#measurements: 13500\\
		$\norm{\Delta x}_2 / \norm{x}_2$ = 5.7\%
}
\label{vanillaC}
        \end{subfigure}
        \hfill
        \begin{subfigure}[b]{0.24\textwidth}
\includegraphics[width=\textwidth]{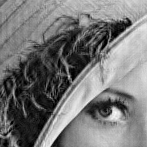} \\
\vspace{-0.07\linewidth}
\includegraphics[width=\textwidth]{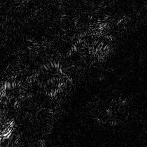} \\
\vspace{-0.07\linewidth}
\includegraphics[width=\textwidth]{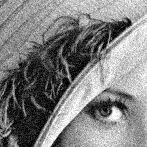}
\caption{
		neighborhood: 24$\times$24 \\
		subsampling: 4$\times$4\\
		\#measurements: 13500 \\
		$\norm{\Delta x}_2 / \norm{x}_2$ = 11.1\%
}
\label{vanillaD}
        \end{subfigure}
\caption{Synthesized images, matched for local covariance maps of a bank of 4 optimized filters (figure~\ref{filters}), are almost indistinguishable from the original. As the neighborhood over which the covariance is estimated increases, the errors increase but are still far less visible than equivalent amounts of additive white noise. Top row: original image $x$ and synthetic images. Middle row: pixelwise magnitude of difference with original $\Delta x$. Each difference image is individually scaled to full dynamic range for display. Bottom row: original image corrupted with additive Gaussian noise, such that the relative error ($\norm{\Delta x}_2/\norm{x}_2$) is the same. %Relative error is the $L_2$ norm of the difference over the $L_2$ norm of the original.
}
\label{vanilla}
\end{figure}
Having optimized a linear transform to reveal the local low-dimensional covariance structure of natural images, we now ask what information these local covariances actually capture about an image. More precisely, we construct a nonlinear representation of an image by filtering it through the learned filter bank, estimating the local covariances and subsampling the resulting covariance map. To explore the power of this representation, we synthesize new images with the same covariance representation. This method of image synthesis is useful for probing
the equivalence class of images that are identical with respect to an analysis model, thereby exhibiting its selectivities and invariances \citep{Portilla:2000gl}.

The procedure for these experiments is as follows. We first build a representation of the original image by estimating the covariance matrix of filter responses in each local neighborhood $P$:
\[ C_P(x) = \left[ \sum\nolimits_{t \in P} w(t) y_i(t) y_j(t) \right]_{i,j}\]
where $w$ is a spatial windowing function. The local covariance map $\phi$ of the original (target) image is then:
\[ \phi( x_{target} ) = \left[ C_P( x_{target} ) \right]_P \]
To synthesize a new image with the same local covariance map, we start with a white noise image $x$, and perform gradient descent on the objective
\[ E(x) = \norm{ \Ve \left( \phi(x) - \phi(x_{target}) \right) }_1 \]
where $\Ve(a)$ is a vector composed of the elements of the multi-dimensional array $a$. We choose an $L_1$ penalty in order to obtain a piecewise quadratic error term, and use a harmonically decaying gradient step to ensure convergence.
\subsection{Perceptual relevance of local covariances}
We distinguish two regimes which lead to very different synthesis results: overcomplete and undercomplete. When $\phi(x)$ is overcomplete, the solution of the synthesis problem at $x = x_{target}$ is often unique. However, even if this holds, finding this solution can be difficult or expensive as the original image must be recovered from a set of quadratic measurements \citep{Bruna:2013vr}. When $\phi(x)$ is undercomplete, many images are represented with the same $\phi(x)$, and synthesis amounts to sampling from this equivalence class of solutions \citep{Portilla:2000gl}.

In an overcomplete setting (figure~\ref{vanillaB}), the simple synthesis algorithm is able to reconstruct the image almost perfectly from the local covariance map. Surprisingly, as we move into the undercomplete regime by further subsampling the covariance map (figure~\ref{vanillaC}), the synthetic images retain excellent perceptual quality.

In the undercomplete regime, the diversity amongst solutions reveals information which is lost by this representation. When we subsample the covariance maps by a factor of 4 in each direction (figure~\ref{vanillaC}), samples include slight phase shifts in high frequency content. When we estimate the covariance over an even larger neighborhood (figure~\ref{vanillaD}), these phase shifts get larger as indicated by the white lines in the difference image (figure~\ref{vanilla}, middle row). Nevertheless, despite the large differences in mean squared error, the synthetic images are almost indistinguishable from the original, especially when compared to images corrupted by additive white noise of equivalent variance (figure~\ref{vanilla}, bottom row).

As a control, we compared these results against syntheses from a representation of natural images in terms of local variance maps. These correspond to a subset of the parameters in local covariance maps, namely the diagonals of the local covariances. To offset the fact that the local variance maps have fewer parameters (the off-diagonal terms of each local covariance being discarded), we increased the number of filters to match the cardinality of the covariance maps. We expected that the results would be similar, as the covariance between two filters can be expressed as the variance of their sum, subtracted by their respective variances. However, the reconstructions from the variance maps are substantially worse (figure~\ref{dimensionality}), both in terms of mean squared error and perceptual quality. This is because successful synthesis from variance maps requires the filters to satisfy a stringent set of properties \citep{Bruna:2013vr}. Synthesis from covariance maps appears to be more robust, both in the oversampled and undersampled regimes.

Having constructed a representation which captures the shape, scale, and orientation of local distributions of filter responses, and tested its perceptual relevance, we now investigate the role of the shape of these distributions.
\begin{figure}
        \begin{subfigure}[b]{0.24\textwidth}
                \includegraphics[width=\textwidth]{pictures/synthesis/vanilla/original.png} \\
\vspace{0.94\linewidth}
\caption{Original image \newline \newline \#pixels: 21600 \newline}
        \end{subfigure}
        \hfill
        \begin{subfigure}[b]{0.24\textwidth}
\includegraphics[width=\textwidth]{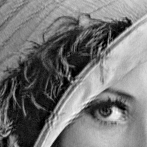} \\
\vspace{-0.07\linewidth}
\includegraphics[width=\textwidth]{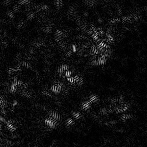}
\caption{
		neighborhood: 8$\times$8 \\
		subsampling: 2$\times$2\\
		\#measurements: 54000\\
		$\norm{\Delta x}_2/\norm{x}_2$ = 8.4\%}
        \end{subfigure}
        \hfill
        \begin{subfigure}[b]{0.24\textwidth}
\includegraphics[width=\textwidth]{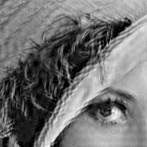} \\
\vspace{-0.07\linewidth}
\includegraphics[width=\textwidth]{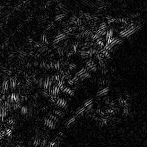}
\caption{neighborhood: 16$\times$16 \\
		subsampling: 4$\times$4\\
		\#measurements: 13500\\
		$\norm{\Delta x}_2/\norm{x}_2$ = 15.4\%
}
        \end{subfigure}
        \hfill
        \begin{subfigure}[b]{0.24\textwidth}
\includegraphics[width=\textwidth]{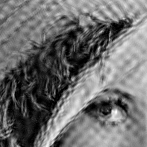} \\
\vspace{-0.07\linewidth}
\includegraphics[width=\textwidth]{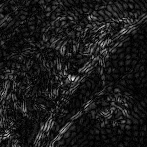}
\caption{neighborhood: 24$\times$24 \\
		subsampling: 4$\times$4\\
		\#measurements: 13500\\
		$\norm{\Delta x}_2/\norm{x}_2$ = 20.7\%}
        \end{subfigure}
\caption{Synthesized images matched for local variance maps fail to capture the relevant structure of the original. Local variances are computed from 10 filters, matching the cardinality of the full covariance representation used in figure~\ref{vanilla}.}
\label{variance}
\end{figure}
\subsection{Perceptual relevance of local low-dimensional covariances}
\begin{figure}
        \begin{subfigure}[b]{0.24\textwidth}
\includegraphics[width=\textwidth]{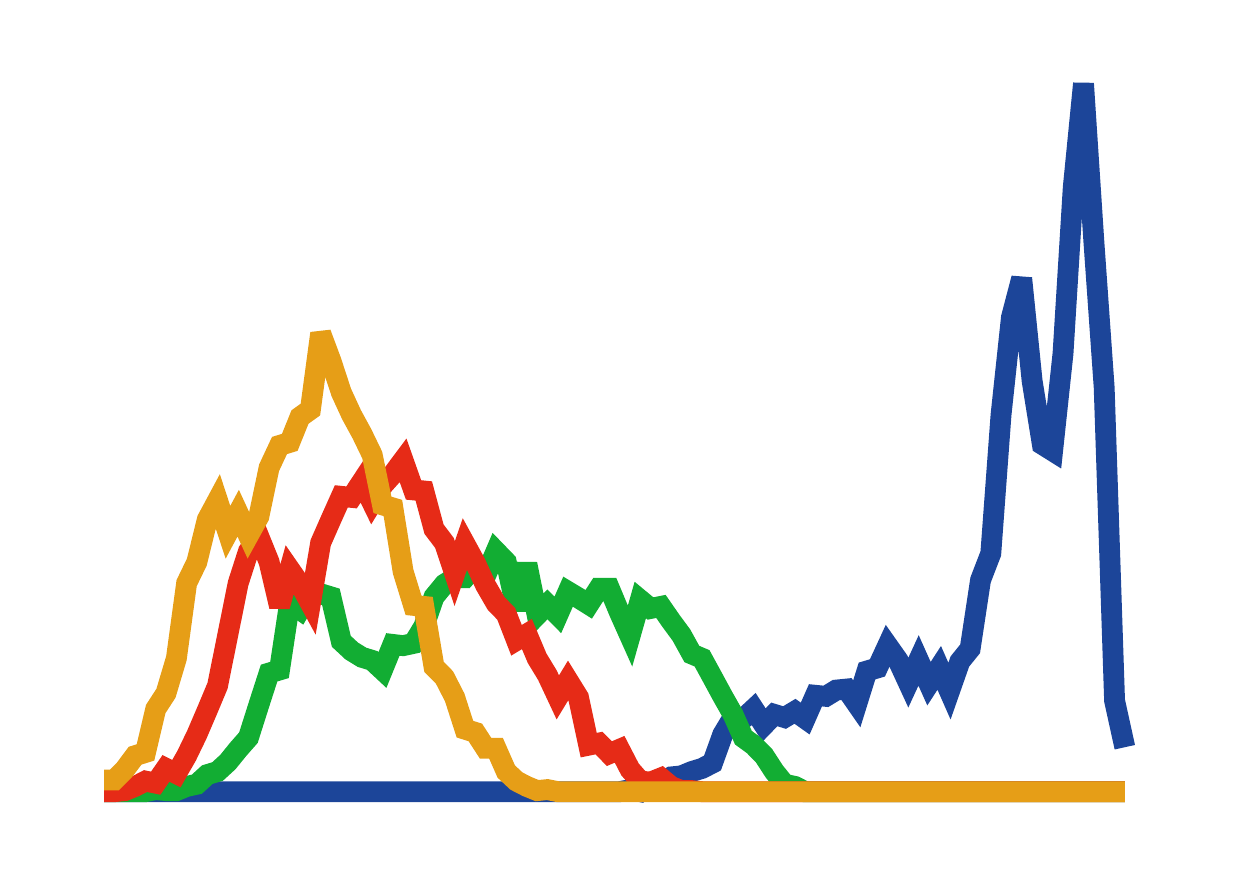}
\vspace{0.1\linewidth}
                \includegraphics[width=\textwidth]{pictures/synthesis/vanilla/original.png} \\
\vspace{0.94\linewidth}
%\caption{Original image \newline \newline \newline Dimensionality: 21600}
        \end{subfigure}
        \hfill
        \begin{subfigure}[b]{0.24\textwidth}
\includegraphics[width=\textwidth]{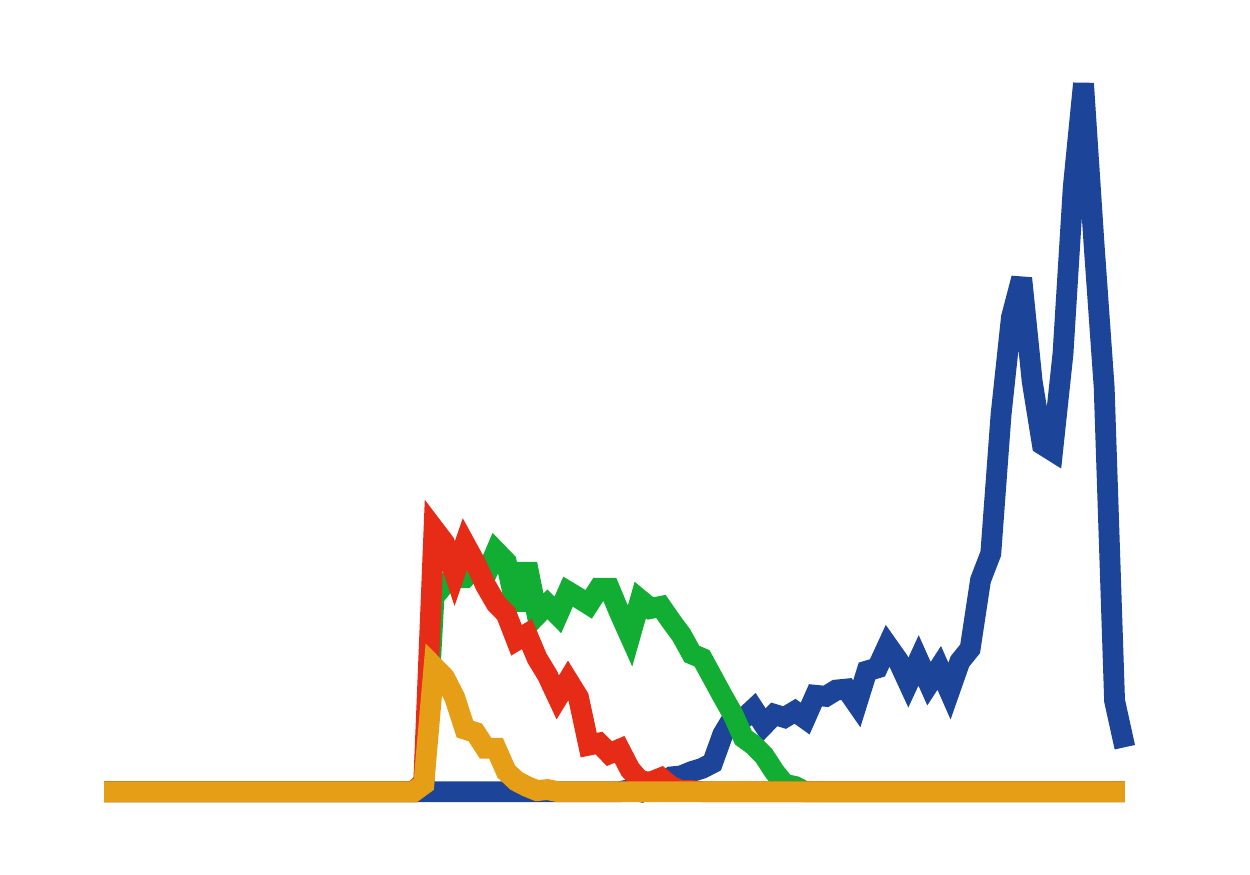}
\vspace{0.1\linewidth}
\includegraphics[width=\textwidth]{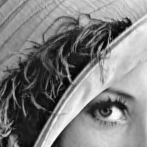} \\
\vspace{-0.07\linewidth}
\includegraphics[width=\textwidth]{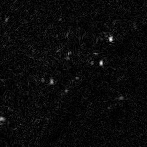}
%\caption{Error = 8.4\% \\ Pooling: 8x8 \\ Subsampling: 2x2\\ \#measurements: 54000}
        \end{subfigure}
        \hfill
        \begin{subfigure}[b]{0.24\textwidth}
\includegraphics[width=\textwidth]{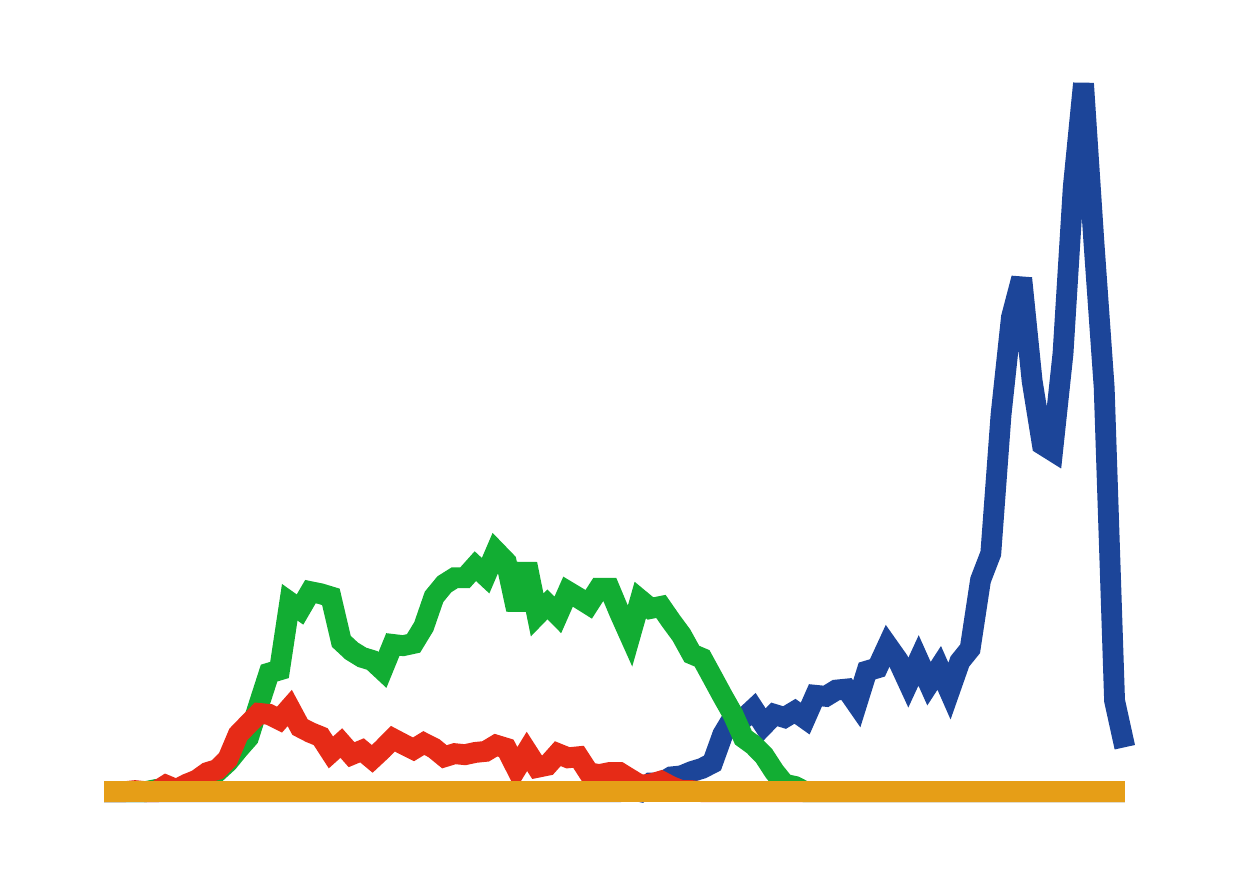}
\vspace{0.1\linewidth}
\includegraphics[width=\textwidth]{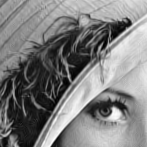} \\
\vspace{-0.07\linewidth}
\includegraphics[width=\textwidth]{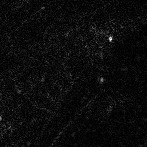}
%\caption{Error = 15.4\% \\ Pooling: 16x16 \\ Subsampling: 4x4\\ \#measurements: 13500}
        \end{subfigure}
        \hfill
        \begin{subfigure}[b]{0.24\textwidth}
\includegraphics[width=\textwidth]{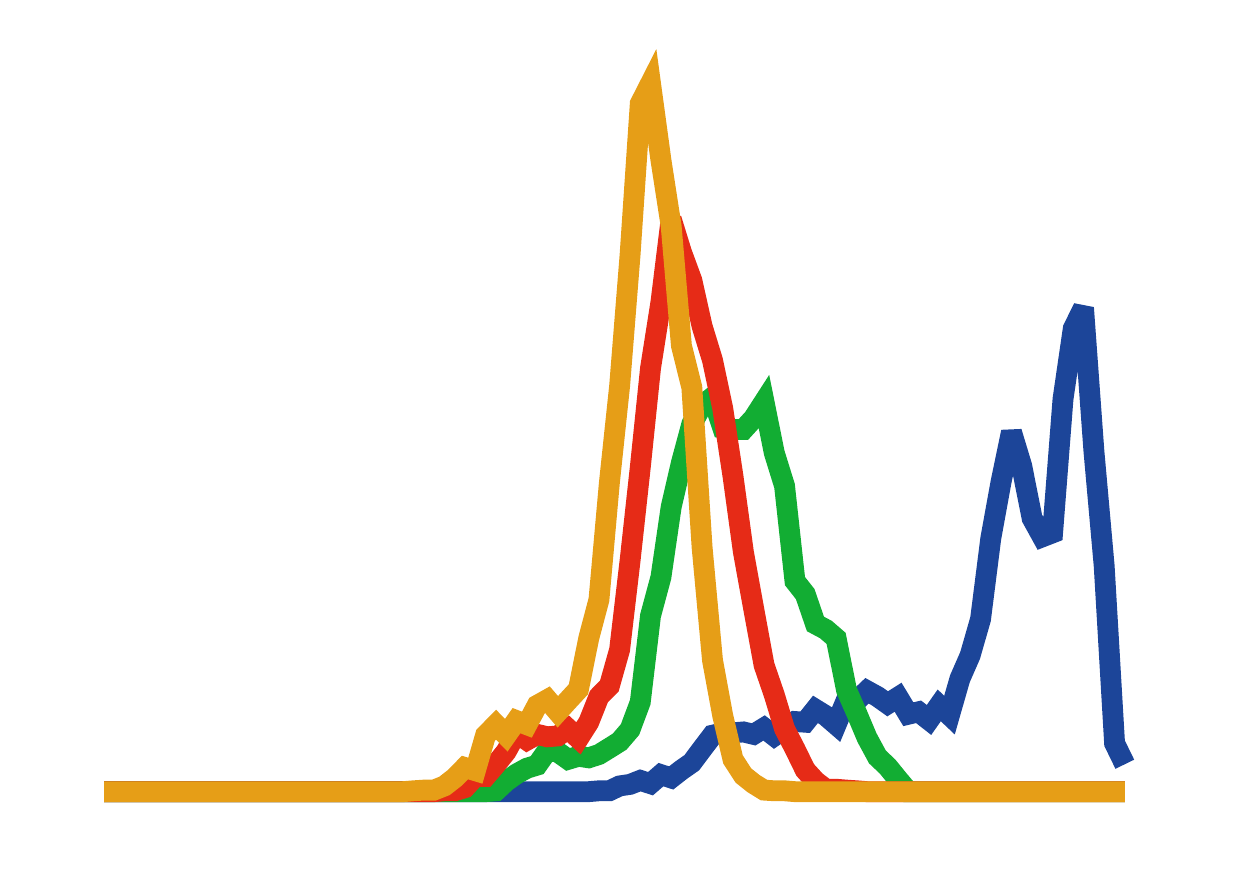}
\vspace{0.1\linewidth}
\includegraphics[width=\textwidth]{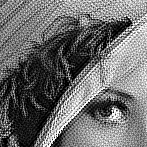} \\
\vspace{-0.07\linewidth}
\includegraphics[width=\textwidth]{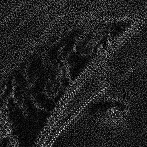}
%\caption{Error = 15.4\% \\ Pooling: 16x16 \\ Subsampling: 4x4\\ \#measurements: 13500}
        \end{subfigure}
\caption{Effects of various modifications of the local covariance map. Top row: histograms of (log) eigenvalues of local covariance matrices. Applying a fixed threshold to the corresponding singular values (2nd column) removes non-oriented content but also low-contrast edges, whereas adaptive thresholding (3rd column) preserves oriented structure regardless of contrast. Dimensionality expansion (4th column) corrupts the image with artificial texture.}
\label{dimensionality}
\end{figure}
In section~\ref{sec:analysis}, we found that natural images distinguish themselves from noise by the proximity of their local distributions to low-dimensional subspaces. We can now ask if these subspaces carry the important information of natural images. Specifically, if we project the representation onto local subspaces, how much of the original image is preserved? We answer this question by synthesizing from a map of local covariances from which we have discarded information corresponding to the smallest eigenvalues. For every covariance matrix, we compute its eigendecomposition, threshold its eigenvalues at a fixed value, and synthesize from the resulting covariance map. Truncating the distribution of eigenvalues results in the removal of high frequency noise as well as low-contrast edges (figure~\ref{dimensionality}, 2nd column). Since a fixed threshold does not distinguish between scale and shape, we repeated the experiment with a threshold value that was scaled adaptively by the local energy (sum of all eigenvalues but the first, which corresponds to the mean luminance). This modifies the shape of local distributions regardless of their scale. Projecting local distributions onto their local subspaces enhances the image by removing noise while preserving any oriented structure (figure~\ref{dimensionality}, 3rd column). On the other hand, making the image locally high-dimensional by raising eigenvalues to a power less than one degrades it by texturizing it artifically (figure~\ref{dimensionality}, 4th column). 

\section{Conclusion}
We have arrived at a representation of natural images in terms of a spatial map of local covariances. We apply a bank of filters to the image, and view the vector of responses at each location as a sample from a multivariate Gaussian distribution with zero mean and a covariance that varies smoothly across space. We optimize the filter bank to minimize the dimensionality of these local covariances; this yields filters that are local and oriented.

We used synthesis to demonstrate the power of this representation. Via a descent method, we impose the covariance map estimated from an original image onto a second, noise image. The resulting image is nearly identical to the original image, and appears to degrade gracefully as we blur and undersample the covariance map. Visually, these reconstructions are vastly better than those obtained from variance maps of equivalent cardinality. The low-dimensionality of the covariances appears to be a crucial factor: when we squeeze the local covariances to make them even more low-dimensional, images retain much of their perceptual quality; when we do the opposite, and make the local covariances more high dimensional, images are corrupted with artificial textures. 

A related line of work by Karklin and Lewicki has studied the statistical fluctuations of natural images. Their initial model describes the variance of filter responses as linear combinations of a sparse set of latent coefficients, thereby approximating the joint distributions of local filter responses as separable \citep{Karklin:2005ct}. More recently, they examined the invariance and discriminability of local filter response distributions, and modeled the log-covariance of image patches as a sparse sum of outer products drawn from a learned dictionary \citep{Karklin:2009hl}. Ultimately, as in traditional sparse models such as sparse coding, ICA, ISA, or K-SVD \citep{Olshausen:1996uo,Bell97,Hyvarinen:2000tk,Rubinstein:2013vs}, these higher order coefficients are assumed to be sparse along the axes of a fixed, finite basis.

On the contrary, the filter responses in our model are not required to be sparse along fixed axes, but along the axes specified adaptively by the eigenvectors of the local covariance matrix. Approximating an arbitrary subspace in a conventional sparse model would require dictionaries of a size that scales exponentially in the dimensionality of the input space (the so-called “curse of dimensionality”). In addition to its computational cost, the high coherence of such an overcomplete dictionary would make the inference of sparse coefficients infeasible with convex relaxations \citep{Donoho:2006ui}. Our model circumvents these problems by continuously parameterizing orientation in feature space. Similarly, continuous parametrizations of translation have been successfully embedded into sparse optimization problems \citep{Ekanadham:2011bb}. These models avoid the brittleness of conventional sparse representations, which exhibit discontinuities when switching coefficients on or off as a signal smoothly varies across space or time. 

%Local low-dimensionality can be interpreted as a kind of sparsity. This is not the traditional sparsity that is defined within the context of fixed, finite bases, as espoused in models such as sparse coding, ICA, ISA, K-SVD, or the Karklin-Lewicki model \citep{Olshausen:1996uo,Bell97,Hyvarinen:2000tk,Rubinstein:2013vs,Karklin:2009hl}. The filter responses in our model are not sparse along the canonical axes, but along the axes specified by the eigenvectors of the local covariance matrix. Thus, it is not the filter responses themselves which are sparse, but the singular value spectra. Conventional notions of sparsity can only achieve this expressive power if one allows their basis to be infinite in size. Furthermore, a locally adaptive sparse basis avoids the brittleness of conventional sparse representations, which exhibit discontinuities when switching coefficients on or off as a signal smoothly varies across space, or over time.

Our model appears promising for a number of image processing applications. The property of local low-dimensionality provides a means of discriminating between natural images and noise, and thus offers a potentially powerful prior for denoising. Our synthesis experiments indicate that undercomplete or thresholded representations of the covariance map are sufficient to reconstruct the original image with a high perceptual quality, suggesting that lossy compression schemes might make use of this representation to produce less visually salient distortions.

Finally, we believe that local covariance map representations offer a natural path for extension into a hierarchical model. As an example, scattering networks have developed decompositions based on alternations of linear filters and local variance maps for applications such as texture synthesis and invariant object recognition \citep{Bruna:2012vu}. Hierarchical models of alternating linear filters and nonlinear pooling have also been proposed as an approximate description of computation along the visual cortical hierarchy in mammals \citep{Riesenhuber99hierarchicalmodels, jarrett-iccv-09}. Our synthesis experiments suggest that a similar infrastructure which recursively stacks linear decompositions and covariance maps, with an objective of reducing local dimensionality, could offer a new canonical description for the biological visual hierarchy, and an unsupervised architecture for use in machine inference, synthesis, and recognition tasks.

\subsubsection*{Acknowledgments}
We would like to thank Joan Bruna for helpful discussions on the use of the nuclear norm and technical aspects of the phase-recovery problem.

\bibliography{iclr2015}
\bibliographystyle{iclr2015}

\end{document}